\documentclass[sn-mathphys]{sn-jnl}
\jyear{2021}%
\theoremstyle{thmstyleone}%
\theoremstyle{thmstyletwo}%
\theoremstyle{thmstylethree}%
\usepackage{subfigure} 
\usepackage{float}
\usepackage[section]{placeins}
\usepackage{appendix}
\usepackage{xpatch}
\usepackage{titletoc}
\usepackage{bm}
\usepackage{ragged2e}
\begin{document}
\xpatchcmd{\thebibliography}{\section*}{\section}{}{}
\title[Re-parameterization Operations Search for Easy-to-Deploy Network]{Efficient Re-parameterization Operations Search for Easy-to-Deploy Network Based on Directional Evolutionary Strategy}

\author[1]{\fnm{Xinyi} \sur{Yu}}

\author[1]{\fnm{Xiaowei} \sur{Wang}}
\email{2112003049@zjut.edu.cn}

\author[1]{\fnm{Jintao} \sur{Rong}}

\author[1]{\fnm{Mingyang} \sur{Zhang}}

\author*[1]{\fnm{Linlin} \sur{Ou}}
\email{linlinou@zjut.edu.cn}

\affil[1]{\orgdiv{College of Information and Engineering}, \orgname{Zhejiang University of Technology}, \orgaddress{\city{Hangzhou}, \postcode{310014}, \country{China}}}

\abstract{ Structural  re-parameterization  (Rep)  methods has achieved significant performance improvement on traditional convolutional network. Most current Rep methods rely on prior knowledge to select the re-parameterization operations. However, the performance of architecture is limited by the type of operations and prior knowledge. To break this restriction, in this work, an improved re-parameterization search space is designed, which including more type of re-parameterization operations.  Concretely, the performance of convolutional networks can be further improved by the search space. To effectively explore this search space, an automatic re-parameterization enhancement strategy is designed  based on neural architecture search (NAS), which can search a excellent re-parameterization architecture. Besides, we visualize the output features of the architecture  to analyze the reasons for the formation of the re-parameterization architecture. On public datasets, we achieve better results. Under the same training conditions as ResNet, we improve the accuracy of ResNet-50 by 1.82\% on ImageNet-1k. }


\maketitle
\section{Introduction}\label{sec1}
Neural architecture search (NAS)\cite{bib14, bib15, bib16, bib17, bib18, bib19, bib20, bib21}  has been widely used in many field, such as object detection\cite{bib0, bib1}, semantic segmentation\cite{bib2,bib3}, image recognition\cite{bib4}, image generation\cite{bib5} and object re-recognition\cite{bib6, bib7}.  Although the architecture searched by NAS is superior to traditional networks in performance such as VGG\cite{bib8}, ResNet\cite{bib9}, MobileNet\cite{bib10,bib11,bib12} and DenseNet\cite{bib13}, its landing ability is far inferior to these networks, which makes it difficult for NAS methods to be widely-used in practice. 

\begin{figure}[b]%
	\centering
	\includegraphics[width=0.7\textwidth]{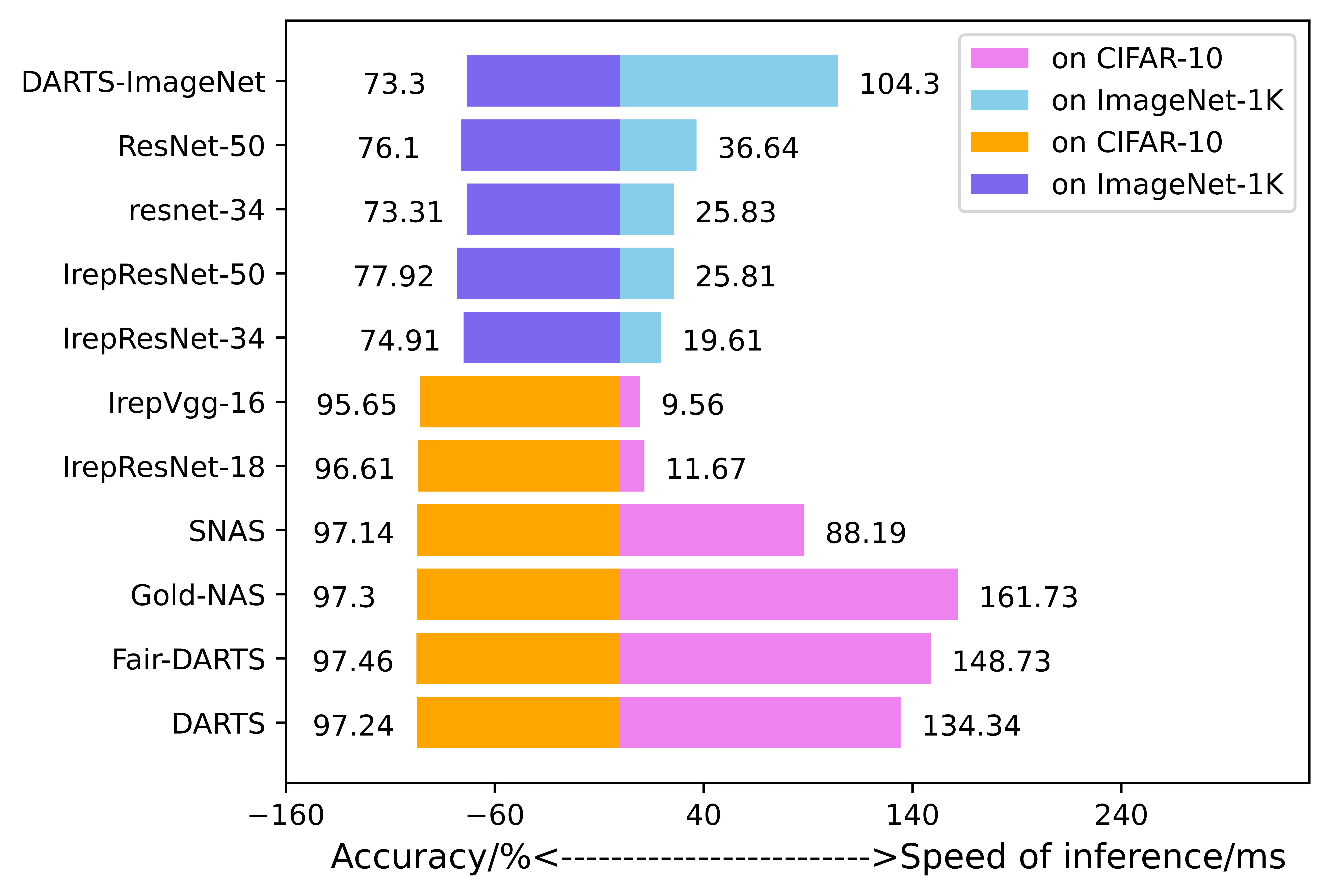}
	\caption{The inference speed and accuracy of networks are tested on NVIDIA AgX Xavier. On ImageNet-1K, our networks achieved better accuracy and inference speed on the general dataset}\label{fig0}
\end{figure}

NAS tends to retain multi-path structures and more "shortcut" operations without artificial restrictions. This structure is not friendly to most terminal devices. Meanwhile, the architecture searched based on the DARTS\cite{bib14} search space achieves better performance on the benchmark dataset, but some convolution operations (such as 7×7 and 5×5 separable convolutions, 7x1--1x7 sequential convolution) in the search space are not well optimized on the terminal devices, which makes them\cite{ bib15,bib16,bib17,bib19, bib44,bib45,bib47} not well deployed on edge devices. In Fig. \ref{fig0}, we tested the accuracy and inference time of networks on the NVIDIA embedded board. The networks searched by traditional search space do not perform well when deployed to edge devices. Especially, the accuracy and inference speed are inferior to the manually designed networks on ImageNet-1K. Therefore, it is a challenging task to improve the performance of the architecture without sacrificing the ability of deployment to terminal devices.

The structural re-parameterization technology provides us with a new idea. Rep methods effectively improve the performance of the tranditional networks, but the specific operations need to be selected based on prior knowledge\cite{bib28, bib29}, such as depth, the type of convolution operations and the number of convolution operations. This approach makes the re-parameterization network suboptimal. To further improve the performance of the model and find the globally optimal networks, we use NAS method to search a set of re-parameterization operations, which can be fused into the traditional networks. In this way, we can get great performance and landing friendly models. In addition, the performance of re-parameterization models is closely related to the type of re-parameterization operations. Some previous work\cite{bib30, bib40} search the best combination of re-parameterization operations automatically, but the type of re-parameterization operations limits the upper limit of network performance. Therefore, the performance of model can be further improved by exploring more re-parameterization operations. 

In addition to the re-parameterization operations  involved in the above work, there are other convolution operations that can be fused into VGG-style networks after appropriate transformation. Thus, an improved re-parameterization search space (IREPS) was designed to make it include more re-parameterization operations, and a better set of re-parameterization operations can be searched from the larger search space, which can be fused into easily deployable networks without accuracy degradation. 

For the larger re-parameterization search space, a directional evolutionary strategy (DES) was designed to explore a optimal architecture population from it. When training the SuperNet parameters, it learns the importance of blocks and candidate operations, and uses them as an indicator to generate different offspring architectures. Thus, the search strategy can directly ignore bad architectures, which makes the algorithm converge rapidly. To explain the reasons for the formation of the architecture and the improvement of performance, finally, we visualized the architecture and the output feature. In summary, the contributions can be summarized as follows:
\begin{enumerate}[1.]
	\item  An improved re-parameterization search space was designed, which contains more re-parameterization operations. Compared with other Rep search space, it can further improve the performance of traditional convolutional network. 
	
	\item To explore the larger search space, DES is proposed to search a set of re-parameterization operations. This search strategy makes a trade-off between the diversity of architectures and the efficiency of search.
	
	\item Extensive experiments on image classification and its downstream tasks demonstrate our architecture achieve better results compared to other related work.
	
\end{enumerate}
\section{Related Work}\label{sec2}
\subsection{Network Architecture Search}\label{subsec2}

Neural architecture search (NAS) is a widely-used technique, which aims to search feature extraction networks that match given tasks. Evolutionary algorithm-based NAS\cite{bib18,bib20,bib21,bib25, bib26} uses the principle of "survival of the fittest" to select architectures and rely on genetics, mutation, crossover and random generation to obtain new offsprings. Evolutionary algorithm is well-established global optimization method with high robustness and wide applicability. However, evolutionary algorithm-based NAS converges slowly due to the random generation of offspring architectures. 

 Gradient-based NAS\cite{bib14, bib15,bib17,bib19,bib34, bib35,bib36, bib37, bib38} benefits from the introduction of differentiable function, which transforms the discrete search space into continuous, so that it can be optimized by gradient optimization algorithm. From the perspective of parameter optimization, it can be divided into two categories. One is bilevel optimization\cite{bib14, bib19, bib35, bib37,bib38}, which optimizes the architecture parameters under the weight parameters are optimal. It can be described as:
\begin{equation}
\min _{\alpha} \mathcal{L}_{v a l}\left(w^{*}(\alpha), \alpha\right)
\textbf { s.t. } w^{*}(\alpha)=\operatorname{argmin}_{w} \mathcal{L}_{\text {train }}(w, \alpha)
\end{equation}
where $\alpha$ denotes architecture, $w_{\alpha}$ denotes the network weight bound with the architecture $\alpha$, $L_{train}$ and $L_{val}$ denote optimization loss on training and validation dataset. It first optimizes the network weights $w$, then finds $\alpha$ that minimizes the validation loss $\mathcal{L}_{v a l}$.
 The other is single-level optimization\cite{bib15, bib36, bib46}, which regards the optimization of $w$ and $\alpha$ as independent processes. It can be described as:
 \begin{equation}
 \alpha^{t}, w^{t}+=\eta \nabla_{\alpha, w} \mathcal{L}_{\text {train }}\left(\alpha^{t-1}, w^{t-1}\right) \label{eq0}
 \end{equation}
Eq. \ref{eq0} indicates that both $w$ and $\alpha$ are optimized in an optimization process. Although gradient-based NAS can converge quickly, the existence of Matthew's effect makes architecture lack of diversity, which leads to the architecture is non-globally optimal. In this work, the single-level optimization approach is used to optimize the weight of SuperNet and learn the importance of operations in the SuperNet. Instead of sampling from SuperNet randomly, DES assumes that the optimal re-parameterization operation combination vary at different training phases (epochs) and aims to generate different architecture based on current optimal re-parameterization operations. Thus, DES can speed up the convergence of search strategy and explore globally optimal architecture.

\subsection{Structural Re-parameterization}\label{subsec2}

The structural re-parameterization technology is an equivalent parameter conversion technology. In our work, the structural re-parameterization technique refers to equivalently converting a multi-branch architecture into a single-branch architecture. ACNet\cite{bib27} proposes to fuse 1D asymmetric convolution into square convolution to enhance the feature representation capability of square convolution. DDB\cite{bib28} aims to enhance the representation of a single convolution by combining diverse branches and give methods for fusing multiple convolution operations in various combinatorial forms. RepVGG\cite{bib29} constructs a residual structure-like branch based on the VGG network and fuses the trained residual-like structure into a 3×3 convolution by structural re-parameterization technique. Based on the re-parameterization technique, RepNAS\cite{bib30} designed a re-parameterization search space, in which all multi-branch structures can be transformed into single-branch structures. There are several re-parameterization techniques, which can be described as: 1) $\textit{Conv-BN}$ to $\textit{Conv}$, 2) a $\textit{Conv}$ for branch addition, 3) $\textit{Sequence Conv}$ structure to $\textit{Conv}$, 4) a $\textit{Conv}$ for depth concatenation to a $\textit{Conv}$, 5) $\textit{K×K average pooling}$ to $\textit{K×K Conv}$, 6) a $\textit{Conv }$for multi-scale $\textit{Convs}$. The above work enhances the feature extraction ability of convolution by re-parameterization technology, but their type of re-parameterization operation is deficient. In this work, we aim to build more different types of reparameterization operations.

\section{Proposed strategy and search space}\label{sec3}
Here, the improved re-parameterized search space is designed first. Secondly, the parameters of the SuperNet are optimized by batch optimization method.  Afterward we introduce how to generate the offspring architectures. Finally, a re-parameterization verification method is implemented to speed up the verification process. 
\subsection{ Improved re-parameterization search space}\label{subsec2}
The convolution operations in traditional convolution networks are called fixed operation. In this work, fixed operation represents $3\times3$ convolution operation.  In the re-parameterization search space, all candidate operations can be fused into fixed operation. Therefore, when the traditional network (ResNet,VGG etc.) that needs to be re-parameterized is determined, the parameters of the architecture is also determined. In addition, when two different operations are re-parameterized, the center weight of the operations need to be aligned , and then fuse the weight parameters. Hence, the re-parameterization search space has the following characteristics:
\begin{enumerate}[1.]
	\item In structural re-parameterization search space, the parameter number of the network is only related to the number of channels. Therefore, changing the number of operations in the block only affects the resource consumption of training, not the amount of parameters and the inference speed when the network is deployed.
	
	\item In the reparameterization search space, the convolution operations with the same groups and channels but different kernel sizes can be fused into each other, if the centers of the convolutions can be exactly overlapped.
\end{enumerate}

In AcNet\cite{bib27}, taking 3×3 convolution as an example, the cruciform weight of the convolution center position has the most important feature information. Thus, the better feature extraction ability can be achieved by enhancing the cruciform feature at the center position of convolution. In this work, more operations are expected to be included in the re-parameterization search space, which can further enhanced the cruciform weight at the center position of the fixed operation. 

Based on the properties of the re-parameterization search space, in this work, $2\times 2, 2\times 1$, and $1\times 2$ dilated convolutions are added to the search space, besides the $3\times 3, 1\times 3, 3\times 1, 1\times 1, 1\times 1-3\times 3$ convolution operations, residual connection and 1×1-average pooling operation. Since the centers of the 1×2 and 2×1 dilated convolutions overlap with the 3×3 convolution, the dilated convolution can be perfectly fused into 3×3 convolution. Specifically, it can be described as $F_{\left ((0) :,:,1:2,::2 \right ) }^{3\times 3} = F_{\left ( D:,:,:,: \right ) }^{1\times 2} $; $F_{\left ((0) :,:,::2,1:2 \right ) }^{3\times 3} = F_{\left ( D:,:,:,: \right ) }^{2\times 1}$;  $F_{\left ((0) :,:,::2,::2 \right ) }^{3\times 3} = F_{\left ( D:,:,:,: \right ) }^{2\times 2}$, where $F_{\left ( 0 \right ) }^{3\times 3} $ represents $3\times 3$ convolution with weights of zero. $F_{\left ( D \right ) }^{1\times 2}$, $F_{\left ( D \right ) }^{2\times 1}$ and $F_{\left ( D \right ) }^{2\times 2}$ represent the weight of $1\times 2, 2\times 1$ and $2\times 2$ dilated convolutions. During re-parameterization, we transfer the weight of dilated convolutions to $3\times3$ convolution with zero weight. In this way, dilated convolutions can be fused into the fixed operation. Considering that ResNet is one of the most widely used models in visual tasks, in this work, to further improve ResNet through a set of re-parameterization operations, we add a re-parameterization block for $3\times3$ convolution that is similar to the residual structure. 

\subsection{Batch optimization of SupNet parameters}\label{subsec3}
In the search process, we use binary encoding $(i.e., {0, 1})$ to cover the SuperNet to obtain different subnets. 1-element indicates participation in forwarding propagation and 0-element indicates non-participation. To trade off efficiency and accuracy, the method of batch optimization of parameters is introduced, which can be expressed as:
\begin{equation}
d \omega=\frac{1}{P} \sum_{i=1}^{P} d \omega_{i}=\frac{1}{P} \sum_{i=1}^{P} \frac{\partial L_{i}}{\partial \omega} \odot \mathcal{M}_{i} \approx \frac{1}{B} \sum_{j=0}^{B} \frac{\partial L_{\mathcal{M}_{j}}}{\partial \omega_{\mathcal{M}_{j}}}  \label{eq1}
\end{equation}
where $P$ and  $B$ represent the number of populations and subnets. $\mathcal{M}$ denotes the subnet sampled from the SuperNet $\mathcal{S}$. $ L_{\mathcal{M}_{j}}$ is the loss value of the subnet on the training dataset. Eq. \ref{eq1} shows that the weight parameters of the SuperNet can be optimized by updating the gradients of subnets in batch. Further, it can be approximated as the average gradients of a part of the individuals in the population.

In search process, the branches number of sub-architectures is limited to $\mathcal{C}$ and all sub-architectures share the weight of SuperNet. The single-level method is used to optimize parameters on the training dataset $D_{train}$. With the formulation used before Eqs. \ref{eq0} and \ref{eq1}, the search process can be given as:
\begin{equation}
\omega_{t+1}, \theta_{t+1}+=\xi_{\omega, \theta}\frac{1}{B} \sum_{j=0}^{B} \frac{\partial \mathcal{L}_{D_{t r a i n}}\left(\omega^{t}, \theta^{t}\right)}{\partial\omega_{\mathcal{M}_{j}},\theta_{\mathcal{M}_{j}}}
\end{equation}
\begin{equation}
\text { s.t. }\left\{\begin{array}{c}\mathcal{M}=\mathcal{S}_{\left\{\mathcal{M}_{1}, \mathcal{M}_{2}, \ldots, \mathcal{M}_{j}\right\}} \\ \left \| {\mathcal{M} }_{j}  \right \|  \leq \mathcal{C}\end{array}\right.  \label{eq5}
\end{equation}
where $\theta$ and $\omega$ are the architecture parameters and the weight parameters of the network, respectively. We generate different sub-architectures to form a population under resource constraints, and optimize the weight and architecture parameters of the Supernet by sampling the architectures from the population.

\subsection{Generation of the architecture}\label{subsec4}
\begin{figure}[t]%
	\centering
	\includegraphics[width=0.9\textwidth]{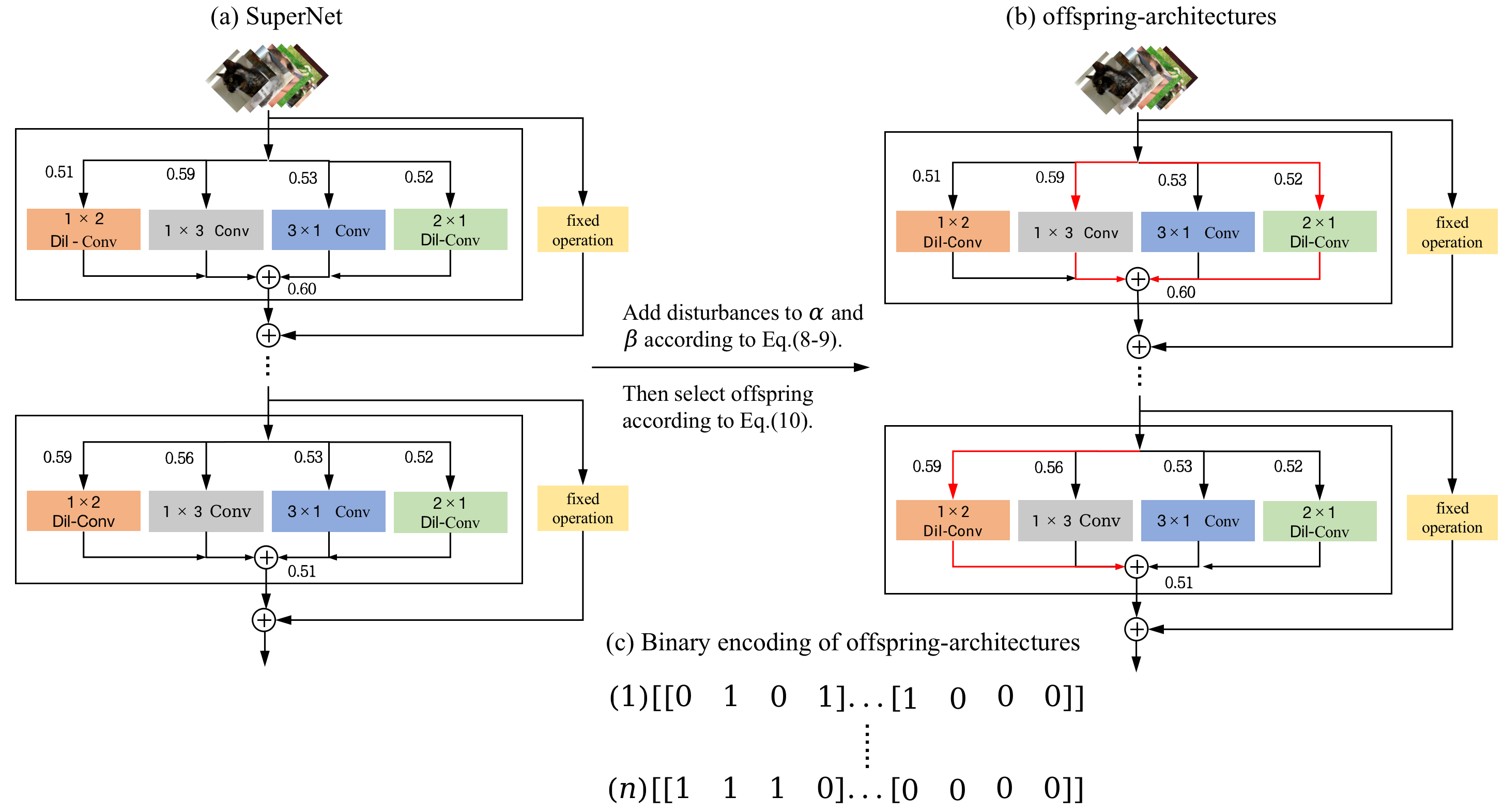}
	\caption{(a): This is the SuperNet. After training it, each branch and block of the Supernet is given different weights. (b): The offspring architectures are generated from the SuperNet according to Eqs. \ref{eq8}-\ref{eq10}. (c): We use binary code to represent the offspring architectures, and the binary code (1) represents the architecture in (b).}\label{fig1}
\end{figure}

The convergence speed of Evolutionary algorithm-based NAS is slow, which is caused by generating offspring architectures randomly. Therefore, we introduce the differentiable method to learn the importance of blocks and re-parameterization operations, and then guide the generation of the sub-architectures. We use the Sigmoid function to quantify the importance of re-parameterization blocks $\beta$ and candidate operations $\alpha$. As shown in Fig. \ref{fig1}, each layer of SuperNet is composed of re-parameterization blocks and fixed operations, the block is composed of multiple candidate operations $O_{p}\left ( \cdot  \right )$. Therefore, the output of the $i^{th}$ layer $\bar{B}^{i}(x)$ can be expressed as:
\begin{equation}
\bar{B}^{i}(x)=\beta_{i}^{\prime} \sum_{o \in O} \frac{1}{1+\mathrm{e}^{-\alpha_{o}^{i}}} f_{o}(x)+F(x) \label{eq6}
\end{equation}
\begin{equation}
\beta_{i}^{\prime}=\frac{1}{1+\mathrm{e}^{-\beta_{i}}}  \label{eq7}
\end{equation}
where $f_{o}(x)$ and $F\left ( x \right ) $ are the output feature of the candidate operations and fixed operation respectively. From Eqs. \ref{eq6} and \ref{eq7}, we can conclude the following easily: 1) From a local perspective, the more significant the enhancement of the fixed operation by the re-parameterization operation in a block, the larger the value of $\alpha$, 2) The feature values of re-parameterization operations are multiplied by the weight $\beta$ of the current block eventually. From a global perspective, the more important the output feature of the block are for the fixed operation, the greater the value of $\beta$. Therefore, when generating offspring architectures, both the global and local characteristics of the architecture are considered.

The architectures in population can be divided into three parts: 1) the architectures retained from the previous population (parent architectures), 2) the offspring architectures generated by the crossover and mutation, 3) the new offspring architectures sampled from the SuperNet. The architectures that are generated according to importance is used to form the third part of the population. Specifically, the random distribution noise $\sigma_{\alpha}$ and $\sigma_{\beta}$ are added to the architecture parameters $\alpha$ and  $\beta$ to ensure the diversity of the architecture. We define $\alpha_{0}^{\prime}=\operatorname{sigmoid}\left(\alpha_{0}\right)$, $\alpha^{\prime}=\operatorname{sigmoid}(\alpha)$, $\beta_{0}^{\prime}=\operatorname{sigmoid}\left(\beta_{0}\right)$, $\beta^{\prime}=\operatorname{sigmoid}(\beta)$, where  $\alpha_{0}$ and $\beta_{0}$  are the initialization weights of $\alpha$ and $\beta$. The range of the perturbation can be defined as:
\begin{equation}
\sigma_{\alpha} \in \left(\alpha_{0}^{\prime}-\max \left(\alpha^{\prime}\right), \alpha_{0}^{\prime}-\min \left(\alpha^{\prime}\right)\right) \label{eq8}
\end{equation}
\begin{equation}
\sigma_{\beta}\in \left(\beta_{0}^{\prime}-\max \left(\beta^{\prime}\right), \beta_{0}^{\prime}-\min \left(\beta^{\prime}\right)\right)\label{eq9}
\end{equation}
where $\sigma_{\alpha}$ and $\sigma_{\beta}$ belong to random distribution. We take the deviation of the maximum and minimum weight values from the baseline as the range of perturbation. When sampling offspring architectures from the SuperNet, the edges with higher weights are retained by global sorting $(\beta_{i}^{\prime}+\sigma_{\beta})\cdot (\alpha_{i}^{\prime}+\sigma_{\alpha})$, $(\cdot)$ denotes the multiplication of two matrices. This process can be simply described as:
\begin{equation}
\left\{\begin{array}{lc}
1, \text { if } \operatorname{rank}\left[\left(\beta_{i}^{\prime}+\sigma_{\beta}\right) \cdot\left(\alpha_{i}^{\prime}+\sigma_{\alpha}\right)\right] \leq \mathcal{C} \\
0, \text { else }
\end{array}\right. \label{eq10}
\end{equation}
where 1-element means the network uses this connection. $rank\left ( \cdot  \right ) $ denotes the global ranking. As shown in Fig. \ref{fig1}, the red line indicates the branch, which selected, and the black line indicates not selected. Due to the addition of  appropriate perturbations, the candidate operations with high weight are retained, and do not completely ignore the operations with low weight in the current stage.

\begin{figure}[b]
	\centering
	\includegraphics[width=0.9\textwidth]{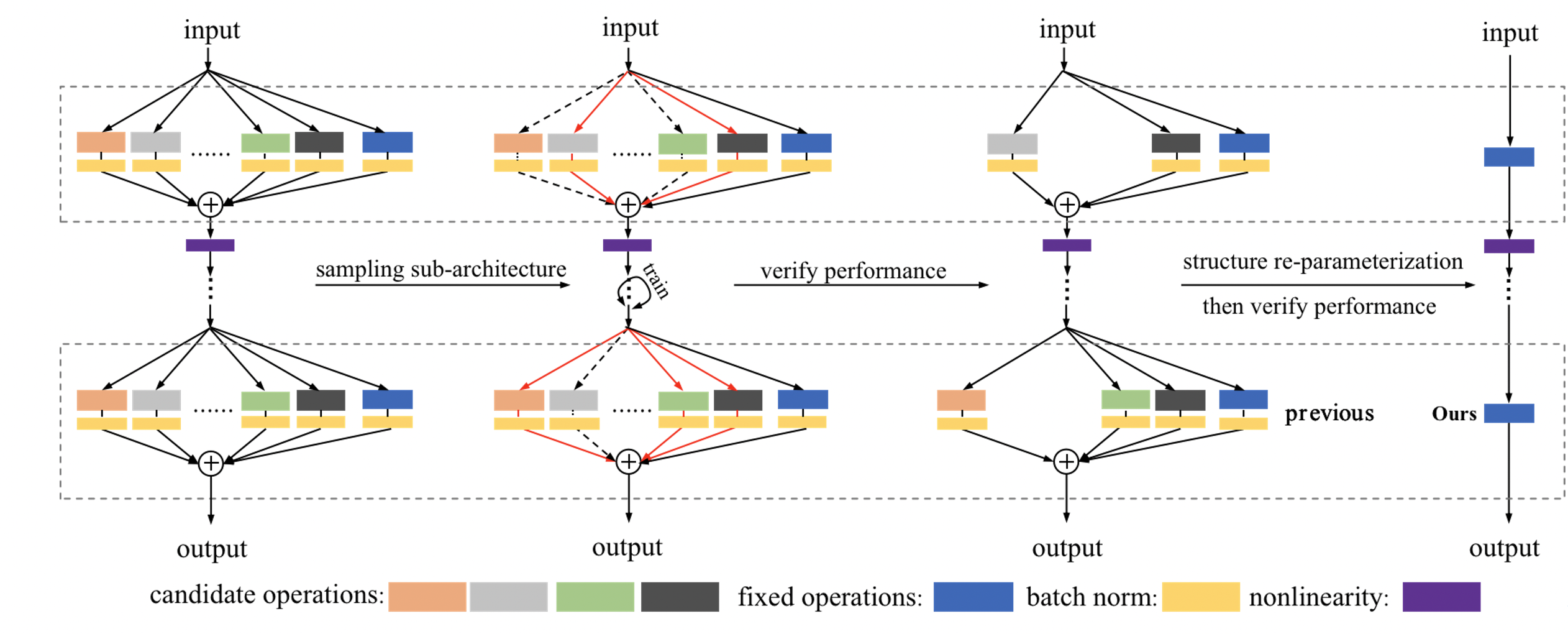}
	\caption{Using the re-parameterized structure to obtain the accuracy. Each sub-architecture in the population can be re-parameterized into the rightmost  structure.}\label{fig3}
\end{figure}

\begin{algorithm}[t]
	\caption{Directional evolution strategy for neural architecture search} \label{algo1}
	\begin{algorithmic}[1]
		\Require SuperNet $\mathcal{S}$, Population $P=\left\{P_{1}, \cdots, P_{k}\right\}$, evolution number $E_{evo} $, Warm up number $E_{warm} $, parameter optimization epochs $E_{p} $, arch-parameters $\alpha$, $\beta$.
		\While{$i< E_{warm}$ do:}
		\State Warm up SuperNet $\mathcal{S}$
		\EndWhile
		\While{$j< E_{evo}$ }
		\While{$k< E_{p}$ }
		\For {Mini-batch data X, target Y in Dataset}
		\State Ramdom sample B sub-architectures from Population.
		\State Forward B sampled sub-networks.
		\State Calculate loss and compute the gradients according to Eq. \ref{eq1}.
		\State Update network parameters $\omega$ and architecture parameters $\alpha$, $\beta$.
		\EndFor
		\EndWhile
		\State Re-parameterize architecture and obtain the performance of the architecture.
		\State Select the architecture according to performance and perform mutation crossover. Meanwhile, sample the subarchitectures from the SperNet according to Eq. \ref{eq8}-\ref{eq10}.
		\EndWhile
		\State Output: Architectures $P=\left \{ P_{1}^{'},  P_{2}^{'} \cdots P_{k}^{'} \right \} $.
	\end{algorithmic}
\end{algorithm}

\subsection{Performance estimation of population}\label{subsec6}
 In evolutionary algorithm-based NAS, evaluating the architectures takes a lot of time. In addition, there is still tiny deviation in the performance of the architecture before and after reparameterization. Although this deviation can be ignored in practical application, we need to search for the best performance architecture after re-parameterization, it has a certain impact on the choice of the architecture. Thus, in this work, the re-parameterization operations are fused into the fixed operation before verifying architecture performance, as shown in Fig. \ref{fig3}.  It is worth noting that the $\textit{BN}$ layer is also fused into the fixed convolution. The multi-branch $\textit{Conv-BN}$ layer becomes $\textit{Conv}$ layer. Hence, using re-parameterized architectures for performance evaluation can speed up the evaluation process and eliminate the deviation.

We use $\alpha$ and $\beta$ to indicate the importance of blocks and candidate operations. Therefore, the weights and biases of the candidate operations need to be scaled $\alpha \cdot \beta $ times before the architecture is re-parameterized. Table \ref{tab1} shows the time consumption of the verification process on CIFAR-10. Concretely, the number of populations is set to [64, 128, 256]. Our approach increases the speed of the architecture evaluation by around $60\%$ compared to the naive evaluation method. In this experiment, considering factors such as architecture diversity search time and computational resources, the population size is set to 128. The overall training procedure is summarized in Algorithm \ref{algo1}.

\section{Experiments}\label{sec4}

To verify the improved re-parameterization search space is effective for different datasets, we search for a set of reparameterization operations for ResNet on CIFAR-10 and ImageNet-1K. Due to ResNet has residual structure, it is natural to weed out residual connection in the search space. Our experiment is divided into two stages: the search stage and the retraining stage.

\begin{table}[b]
	\begin{center}
		\renewcommand{\raggedright}{\leftskip=0pt \rightskip=0pt plus 0cm}
		\caption{We obtain the performance of architecture on Nvidia A100 GPU. The time to evaluate the performance of the architecture by the re-parameterization techniques consists of two parts, i.e. the time consumed by the architecture re-parameterization and the architecture forward inference. The results are average of verifying 10 populations and the batch size is 512, full precision(fp32).}\label{tab1}
		\resizebox{\linewidth}{!}{
			\begin{tabular}{cccccc}
				\toprule
				\multicolumn{2}{c}{Population Size/Time}                   & 64/S   & 128/S  & 256/S  \\ 	\midrule
				\multirow{2}{*}{Multi-Branch}                   & VGG-16    & 1349.4 & 2681.1 & 5336.9 \\
				& Resnet-18 & 649.1  & 1301.6 & 2685.7 \\ \hline
				\multirow{2}{*}{Re-parameterized}               & VGG16    & 499.5  & 1000.0 & 1823.9 \\
				& Resnet-18 & 286.3  & 571.5  & 1133.9 \\ \hline
				\multirow{2}{*}{Acceleration   percentage (\%)} & VGG-16    & 63.7   & 62.7   & 65.8   \\
				& Resnet-18 & 55.9   & 56.1   & 57.8   \\  \botrule
			\end{tabular}
		}
	\end{center}
\end{table}

\subsection{Search Architectures on CIFAR-10}\label{subsec2}
We add a re-parameterization block similar to the residual structure to the $3 \times3$ convolution in VGG-16 and ResNet-18. For a fair comparison, the structure of the network and data augmentation techniques are followed by ACNet\cite{bib27} and ResNet\cite{bib9}. We use the SGD optimizer with a learning rate of $0.1$ to optimize the parameters of the network. To optimize the architecture parameters $\alpha$ and $\beta$, we use Adam optimizer with learning rate of $0.0001$ and $\left ( 0.5, 0.999 \right )$ betas. We limit the number of branches to $\frac{2}{3}$ times of the total branch number. In the training process, we sampled $5$ architectures and used them to update SuperNet. The probability of both mutation and crossover for the architecture is 0.5. 

We search for 500 epochs and retrain architectures on CIFAR-10 dataset. Except for the learning rate and the probability of the drop-path, the retraining process are the same as DARTS\cite{bib14}. Respectively, the learning rate and the probability of drop-path is set to 0.05 and 0.08.

Table \ref{tab2} shows our results. We achieved $1.02\%$ better accuracy than RepVGG\cite{bib29} and $0.21\%$ than RepNAS\cite{bib30}. Our architecture has a great advantage in the inference process. Since the re-parameterized architecture retains only convolution and non-linear operations, the inference speed of IrepResNet-18 and IrepVGG-16 reaches 3.93ms and 1.76ms per image, which is faster than the architectures such as DARTS.

\begin{table}[b]
	\begin{center}
		\renewcommand{\raggedright}{\leftskip=0pt \rightskip=0pt plus 0cm}
		\caption{Comparison with state-of-the-art image classifiers on CIFAR-10 dataset. We calculated the parameters of the model and tested the model of the inference time on Nvidia A100 GPU with a batch size of 1, full precision(fp32).}\label{tab2}
		\resizebox{\linewidth}{!}{
			\begin{tabular}{cccccc}
				\toprule
				Model          & Top-1 (\%) & Params(M) & Inference (ms) & Search Cost (GPU   days)  \\ \midrule
				VGG            & 94.12      & 14.73     & 2.14           & —                                  \\ 
				ResNet-18      & 96.21      & 11.69     & 4.12           & —                                   \\ 
				RepVGG         & 94.62      & 14.73     & 1.76           & —                                      \\ 
				RepNAS (VGG)   & 95.43      & 11.69     & 1.76           & 0.7                 \\ 
				AcNet          & 94.47      & 14.73     & 1.76           & —                                  \\  \hline
				DARTS (second) & 97.24      & 3.3       & 31             & 4                         \\ 
				P-DARTS        & 97.50       & 3.4       & 33             & 0.3                       \\ 
				GoldNAS        & 97.39      & 3.67      & 40             & 1.1                        \\ 
				CARS           & 97.38      & 3.6       & 27             & 0.4                           \\ 
				AmoebaNet-A    & 96.60      & 3.2       & 38             & 3150                  \\ \hline
				IrepResNet-18  & \textbf{96.61}      & \textbf{11.69}     & \textbf{3.93}           & \textbf{8.0}                     \\ 
				IrepVgg-16     & \textbf{95.65}      & \textbf{14.73}     & \textbf{1.76}           & \textbf{16.0}                      \\ 
				\botrule
			\end{tabular}
		}
	\end{center}
\end{table}

\subsection{Experience on ImageNet-1K}\label{subsec3}
To reveal the generalization ability, we evaluate on the ImageNet-1K, which contains 1.3M images for training and 50K for validation from 1000 classes. To save computational resource and speed up the search, based on the conclusion of ACNet\cite{bib27}, we remove the $2\times 2$ dilated convolution from the search space. We set $B=1$, $E_{warm}=5$ and batch size is 256. We use Adam optimizer with 0.0001 learning rate and (0.5, 0.999) betas to optimize $\alpha$, $\beta$. We limit the number of branches to $\frac{1}{2}$ times of the total branch number. We search for 100 epochs and then fix the structure of the SuperNet to retrain architectures for 120 epochs. To be fair, we use the same data augmentation techniques as ResNet\cite{bib9}. 

\begin{table}[t]
	\begin{center}
		\renewcommand{\raggedright}{\leftskip=0pt \rightskip=0pt plus 0cm}
		\caption{Results of our models on ImageNet-1K dataset compared to other models. The performance on ImageNet-1K with comparison to other NAS methods and models. All experiments on the ImageNet-1K were performed based on Nvidia A100 GPU. We calculated the parameters of the model and tested the model of the inference time with a batch size of 1, full precision(fp32).}\label{tab3}
		\resizebox{\linewidth}{!}{
			\begin{tabular}{cccccc}
				\toprule
				Model                                                                 & Top-1(\%) & Top-5(\%) & Params(M) & Inference   (ms) & Search   Cost (GPU-days) \\  \midrule
				ResNet18                                                              & 69.76     & 89.07     & 11.69     & 4.25             & —                        \\
				ResNet34                                                              & 73.31     & 91.42     & 21.80     & 6.12             & —                        \\
				ResNet50                                                              & 76.10     & 93.29     & 25.56     & 7.54             & —                        \\ \hline
				\begin{tabular}[c]{@{}c@{}}DyRep \\ (ResNet-18)\end{tabular}     & 71.58     & —         & 16.90     & 3.59             & —                        \\
				\begin{tabular}[c]{@{}c@{}}DyRep  \\ (ResNet-34)\end{tabular}     & 74.68     & —         & 33.10     & 5.15             & —                        \\
				\begin{tabular}[c]{@{}c@{}}DyRep \\ (ResNet50)\end{tabular}      & 77.08     & —         & 31.50     & 6.14             & —                        \\ \hline
				\begin{tabular}[c]{@{}c@{}}DDB  \\ (ResNet18)\end{tabular}        & 70.99     & —         & 26.30     & 3.59             & —                        \\
				\begin{tabular}[c]{@{}c@{}}DDB \\ (ResNet34)\end{tabular}        & 74.33     & —         & 49.90     & 5.15             & —                        \\
				\begin{tabular}[c]{@{}c@{}}DDB \\ (ResNet50)\end{tabular}        & 76.71     & —         & 40.70     & 6.14             & —                        \\ \hline
				DARTS   (second)                                                      & 73.30     & 91.3      & 4.70      & 67.4             & 4                        \\
				P-DARTS                                                               & 75.60     & 92.6      & 4.90      & 62.3             & 0.3                      \\
				GoldNAS                                                               & 76.10     & 92.7      & 6.40      & 39.4             & 1.7                      \\
				CARS                                                                  & 75.20     & 92.5      & 5.10      & 59.9             & 0.4                      \\ \hline
				\begin{tabular}[c]{@{}c@{}}IrepResNet-18\end{tabular} & \textbf{71.57}     & \textbf{89.98}     & \textbf{18.23}     & \textbf{3.59}             & \textbf{12}                       \\
				\begin{tabular}[c]{@{}c@{}} IrepResNet-34\end{tabular} & \textbf{74.91}     & \textbf{92.12}     & \textbf{36.12}     & \textbf{5.15}             & \textbf{19}                       \\
				\begin{tabular}[c]{@{}c@{}} IrepResNet-50\end{tabular} &\textbf{77.92}     & \textbf{93.88}     & \textbf{29.04}     &\textbf{6.14}             & \textbf{30}                       \\ \botrule
			\end{tabular}
		}
	\end{center}
\end{table}

We compare our architectures with state-of-the-arts in Table \ref{tab3}. Compared to other work, IrepResNet also shows favorable performance. IrepResNet-50 achieve top-1 accuracy of 77.92\%, which is 1.82\% higher than ResNet-50, 0.84\% higher than DyRep\cite{bib40} and 1.21\% higher than DDB\cite{bib28}. IrepResNet-34 and IrepResNet-18 also achieve great performance. Meanwhile, the architecture has faster inference speed compared to ResNet, DARTS, P-DARTS, GoldNAS, etc.

\begin{figure}[t]
	\centering
	\includegraphics[width=1\textwidth]{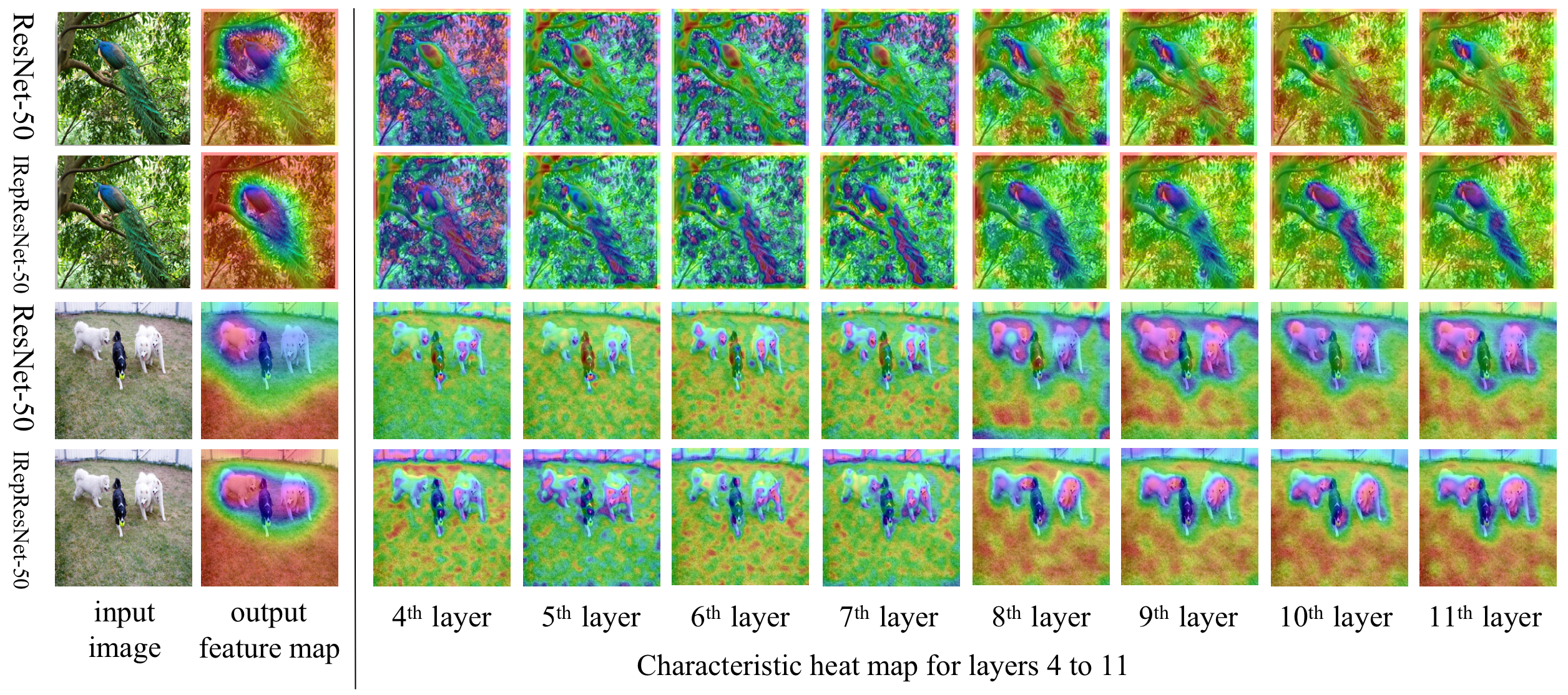}
	\caption{We visualized the output feature values of the convolution in the ResNet-50 and IrepResNet-50 to better interpret the structure of our network. We consider the structure of Conv1×1-Conv3×3-Conv1×1 as a layer. The first and second columns are the original image and the heatmap of the last layer output. The other columns are the heatmap of the feature outputs convolved in the 4th layer to the 11th layer. IRepResNet-50 has significantly better feature focus than ResNet-50.}\label{fig4}
\end{figure}

We plot the structure of IrepResNet in Appendix \ref{secA1}. The structure of IrepResNet-50 is truncated in the middle, i.e., the first eight layers retain all re-parameterization operations, and the last eight layers exclude all enhancement operations. To better explain this phenomenon, we visualized the output feature values of ResNet-50 and IrepResNet-50. As shown in Fig. \ref{fig4}, it can be easily concluded that IrepResNet-50 has stronger discrimination ability for targets compared to ResNet-50. For ResNet-50, the role of the first eight layers task is mainly to achieve the separation of foreground and background in the image. While the last eight layers task is mainly to further distinguish the subtle differences between foreground and background, so that the target in the image can be focused accurately. This division of task is significant for the formation of the IrepResNet-50 structure. The first eight layers obtain less feature  information
due to the few number of channels. Thus, the re-parameterization operations are important to improve the current feature information, which makes the them has higher weight. The last eight layers can acquire more feature information because of the more channels, which allows it to take on the task of focusing on the target and refining the foreground and background in the picture. Compared with the first eight re-parameterization blocks, the feature information in the last eight re-parameterization blocks is not important for the last eight fixed operations, which makes the re-parameterization operations has smaller weight value. Therefore, searching the architecture under resource constraints makes the algorithm prefer to retain the re-parameterization operations that in the shallow layers. We also visualized the output features and architecture of IrepResNet-34 and IrepResNet-18, as shown in Appendix \ref{secA1}. The 
re-parameterization operations that are preserved in the IrepResNet-34 and IrepResNet-18 also emerge the same trend. 

\subsection{Generalization performance on Downstream Task} \label{subsec4}
We transfer our ImageNet-pretrained IrepResNet-50 and IrepResNet-18 model to downstream tasks object detection to validate generalization of the model. Specifically, the pre-trained model is used as the backbone for the downstream algorithms FPN\cite{bib41} and CenterNet\cite{bib42} algorithms on the COCO dataset. For the optimization of the target detection model, we refer to the optimization approach and hyperparameter settings of MMDetection\cite{bib43}. FPN and CenterNet are fine-tuned on a single NVIDIA A100 GPU with batch sizes 16 and 64, respectively. In addition, the fine-tuned model can re-parameterize the backbone to achieve faster forward inference. The results in Table \ref{tab4} show that IrepResNet can achieve better performance compared to FPN, CenterNet, and DyRep.
\begin{table}[t]
		\begin{center}
			\caption{Results on object detection.}\label{tab4}
			\resizebox{\linewidth}{!}{
				\begin{tabular}{cccc}
					\toprule
					Backbone       & Algorithm & \begin{tabular}[c]{@{}c@{}}ImageNet  Top-1\end{tabular} & \begin{tabular}[c]{@{}c@{}}COCO mAP\end{tabular} \\ \midrule
					ResNet-18      & CenterNet & 69.76                                                       & 29.5                                                  \\
					ResNet-50      & FPN       & 76.10                                                       & 37.9                                                  \\
					DyRep-ResNet50 & FPN       & 77.08                                                       & 38.1                                                  \\ \hline
					Irep-ResNet18  & CenterNet & 71.27                                                       & \textbf{31.2}                                                  \\
					Irep-ResNet50  & FPN       & 77.92                                                       & \textbf{38.2}                                                  \\ \botrule
			\end{tabular}
		}
	\end{center}
\end{table}
\begin{table}[t]
	\begin{center}
		\renewcommand{\raggedright}{\leftskip=0pt \rightskip=0pt plus 0cm}
		\caption{Performance of the architecture on the CIFAR-10 dataset under different resource constraints.}\label{tab5}
		\resizebox{\linewidth}{!}{
			\begin{tabular}{ccccccc}
				\toprule
				Model         & 1/6   & 1/3   & 1/2   & 2/3   & 5/6   & 1     \\ \midrule
				IRepVGG-16    & 94.22 & 94.58 & 95.21 & 95.65 & 95.64 & 95.64 \\
				IRepResNet-18 & 96.28 & 96.47 & 96.47 & 96.61 & 95.50 & 96.52 \\ \botrule
			\end{tabular}
		}
	\end{center}
\end{table}
\begin{figure}[t]%
	\centering
	\includegraphics[width=0.7\textwidth]{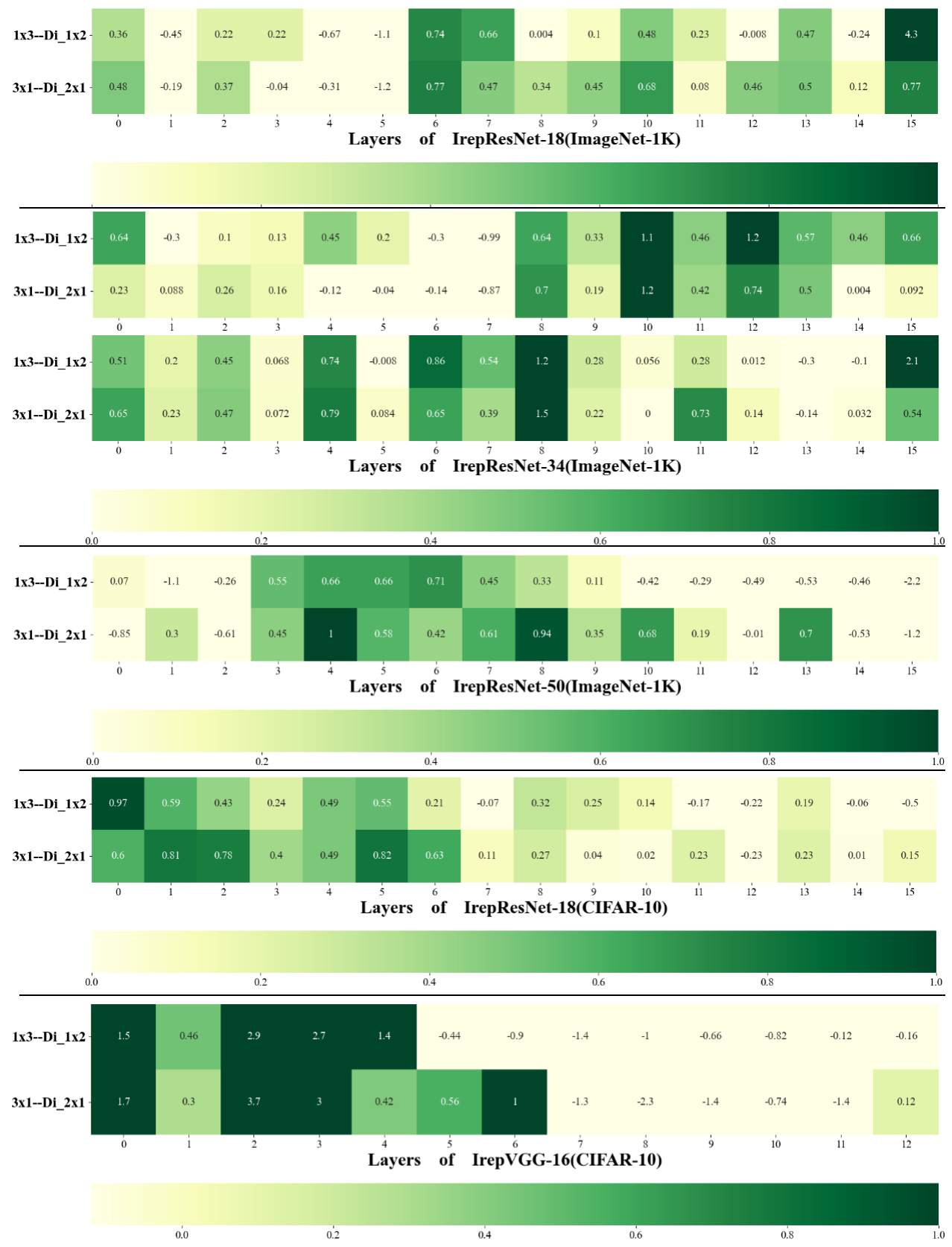}
	\caption{Differential heatmap of the $1\times 3$ convolution--$1\times 2$ dilated convolution and $3\times1$ convolution--$2\times1$ dilated convolution, which searched on CIFAR-10 and ImageNet-1K dataset}\label{fig6}
\end{figure}
\subsection{Ablatilaon Study}\label{subsec5}
\subsubsection{Search under different resource constraints}\label{subsec3}
To explore the impact on the performance of the architecture under different resource constraints, we searched IrepVGG-16 and IrepResNet-18 on the CIFAR-10 dataset. Specifically, the number of branches is set to $\left [ \frac{1}{6},\frac{1}{3},\frac{1}{2}, \frac{2}{3},\frac{5}{6},1  \right ] $ times of the total branch number. As shown in Table \ref{tab5}, when the resource constraint reaches
 2/3, the architecture achieves better performance. We found that the performance of the architecture is weaker than RepNAS\cite{bib30} when retaining the same number of branches as RepNAS in the improved reparameterization search space (retain four branches for each block). Based on AcNet\cite{bib27}, the main reason is that the enhancement effect of the dilated convolutions on the $3\times 3$ convolution is weaker than the $1\times 3$ and $3\times1$ convolutions. As shown in Fig. \ref{fig6}, the architecture weights of the $1\times 3$ convolution--$1\times 2$ dilated convolution and $3\times1$ convolution--$2\times1$ dilated convolution are subtracted and transformed equivalently and we plotted them as heatmap. The larger the difference values, the more important the asymmetric convolution ($3\times1$ and $1\times3$ convolutions) in the same layer. This indicates that the feature enhancement effect of asymmetric convolutions ($1\times 3$ and $3\times1$ convolution) are stronger than dilated convolutions in this experiment. Therefore, when the same number of branches are retained as RepNAS\cite{bib30}, some of the $1\times 3$ and $3\times1$ convolutions may be replaced by dilated convolution due to the Matthew effect of the gradient-based learning method, which leads to the architectures with potentially weaker performance than RepNAS\cite{bib30}. 

\section{Conclusion}\label{sec5}
In order to further improve tranditional convolutional networks, we designed a more comprehensive re-parameterization search space and searched it by directional evolutionary strategy to further improve the performance of ResNet. A re-parameterization block similar to the residual connection is added to each $3\times3$ convolution in ResNet model. Then finding an optimal architecture population by exploring the derivative architectures of the optimal re-parameterization architecture at the current stage. Extensive experiments demonstrate that the proposed the improved re-parameterization search space can further improve the performance of models and perform well in downstream tasks. Moreover, we explain the reasons for the formation of the architecture and analyze the enhancement effect between re-parameterization operations. It is worth mentioning that the improved re-parameterization search space proposed in our paper can be further used as a bridge between coarse-grained search and fine-grained search. It means that the re-parameterization model after coarse-grained search (architecture operation) can be divided into $1\times1$ convolution and $2\times 2, 2\times 1, 1\times 2$  dilated convolution, and then the model can be transformed into a channel pruning friendly network, which can actually further reduce the FLOPs and inference time of the model.
\\
\\
\begin{appendices}

\section{IrepResNet model}\label{secA1}
		\begin{figure}[H]
			\centering
			\includegraphics[width=0.86\textwidth]{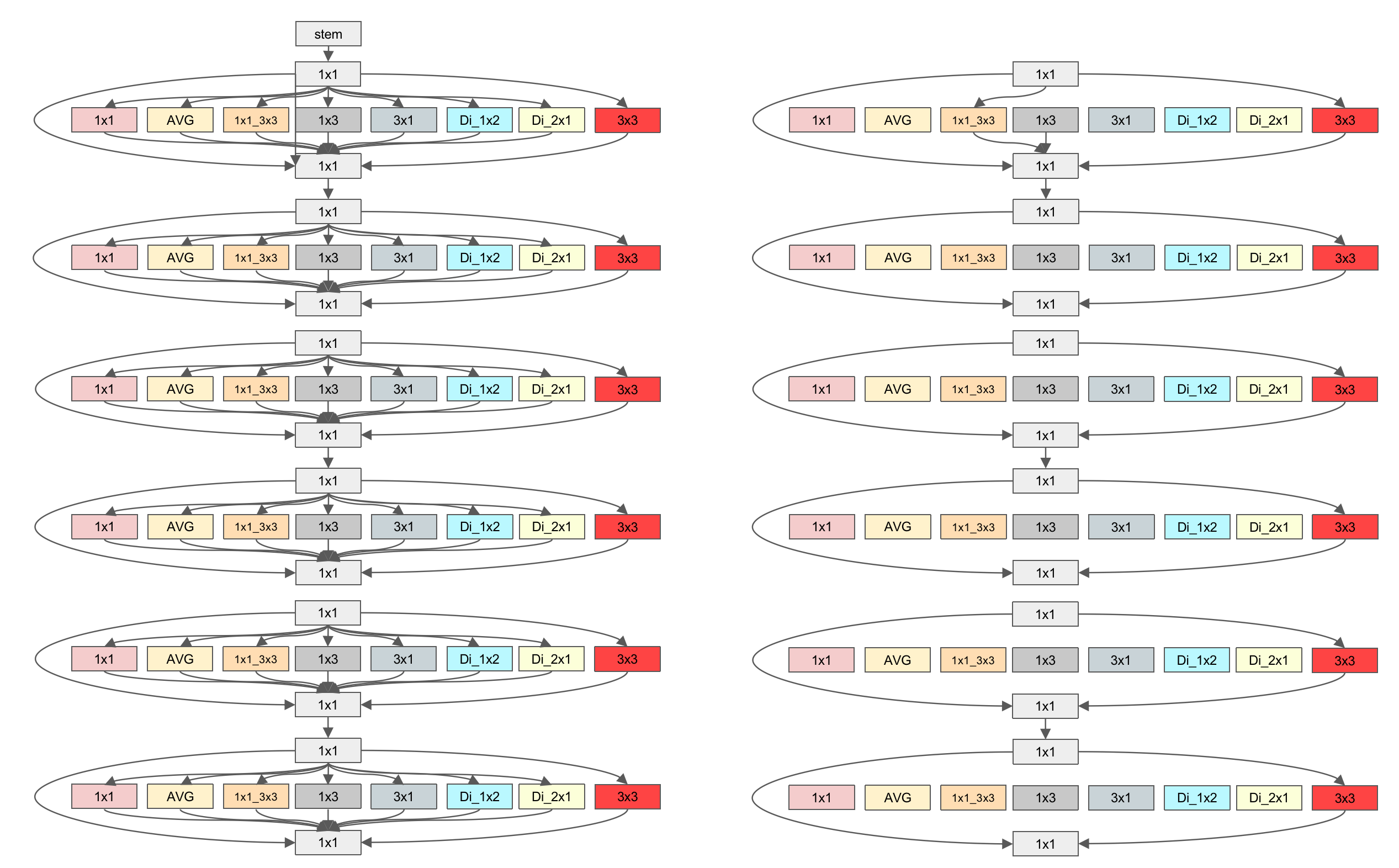}
			\includegraphics[width=0.86\textwidth]{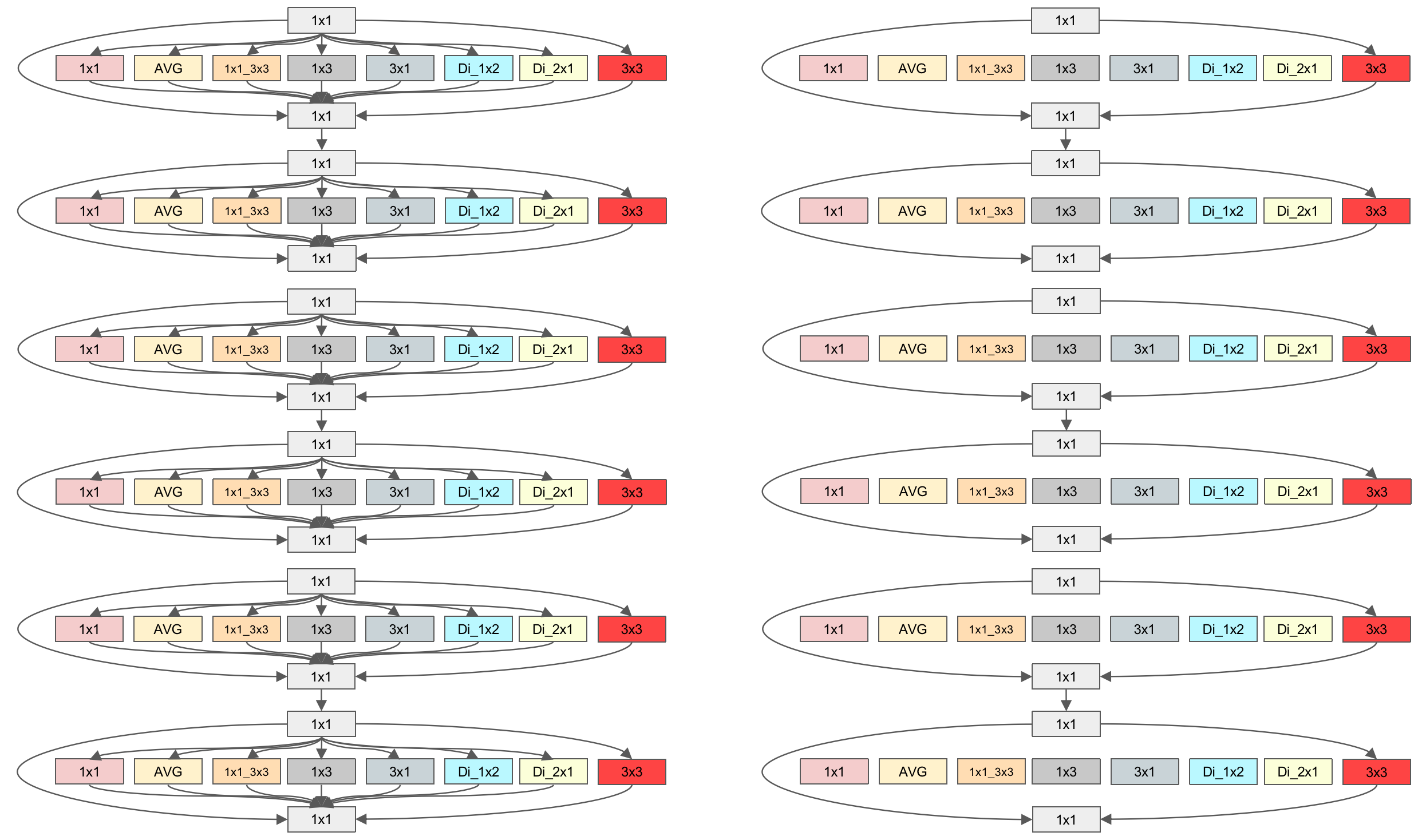}
			\includegraphics[width=0.86\textwidth]{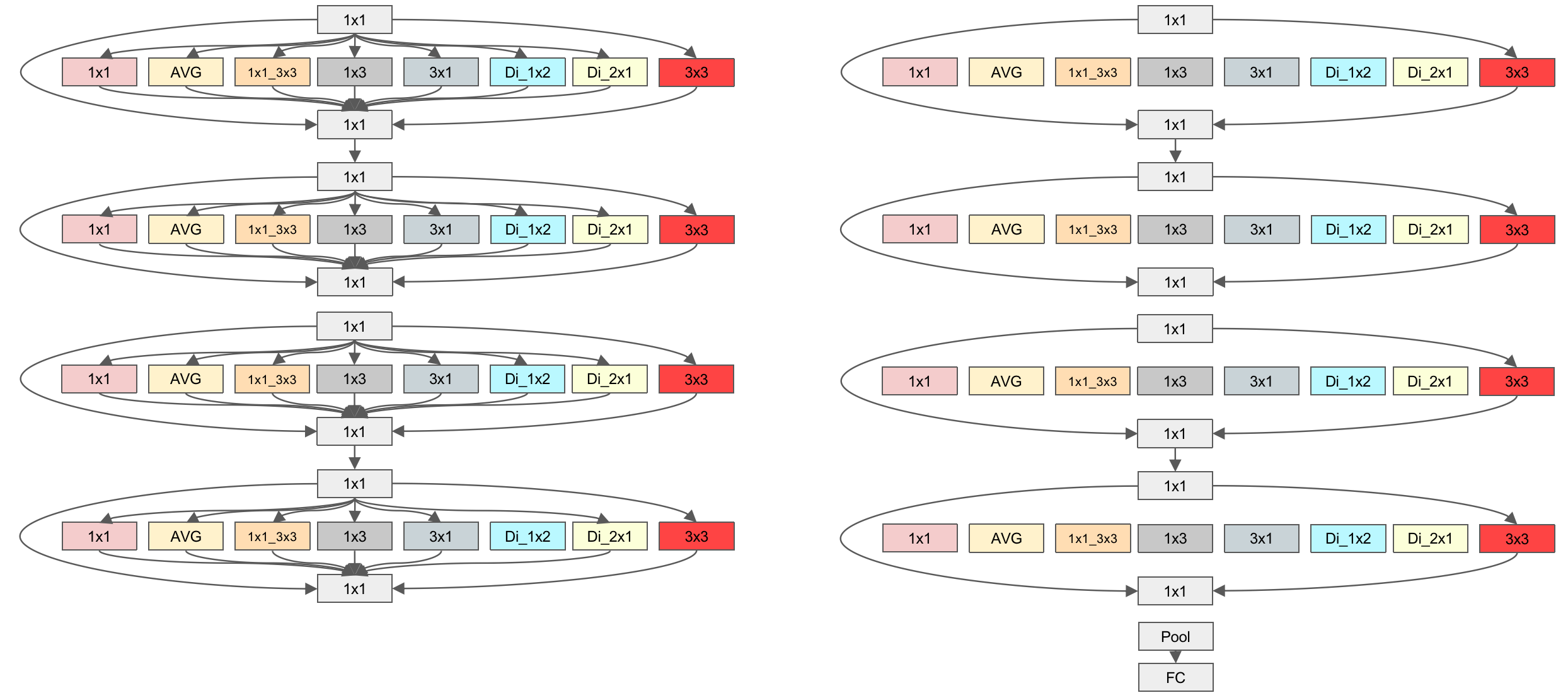}
			\caption{IrepResnet-50 searched on ImgeNet. The $3\times 3$ convolution operation is a fixed operation and does not participate in the search process of the architecture.}\label{fig7}
		\end{figure}
	\begin{figure}[H]
		\centering
		\includegraphics[width=0.85\textwidth]{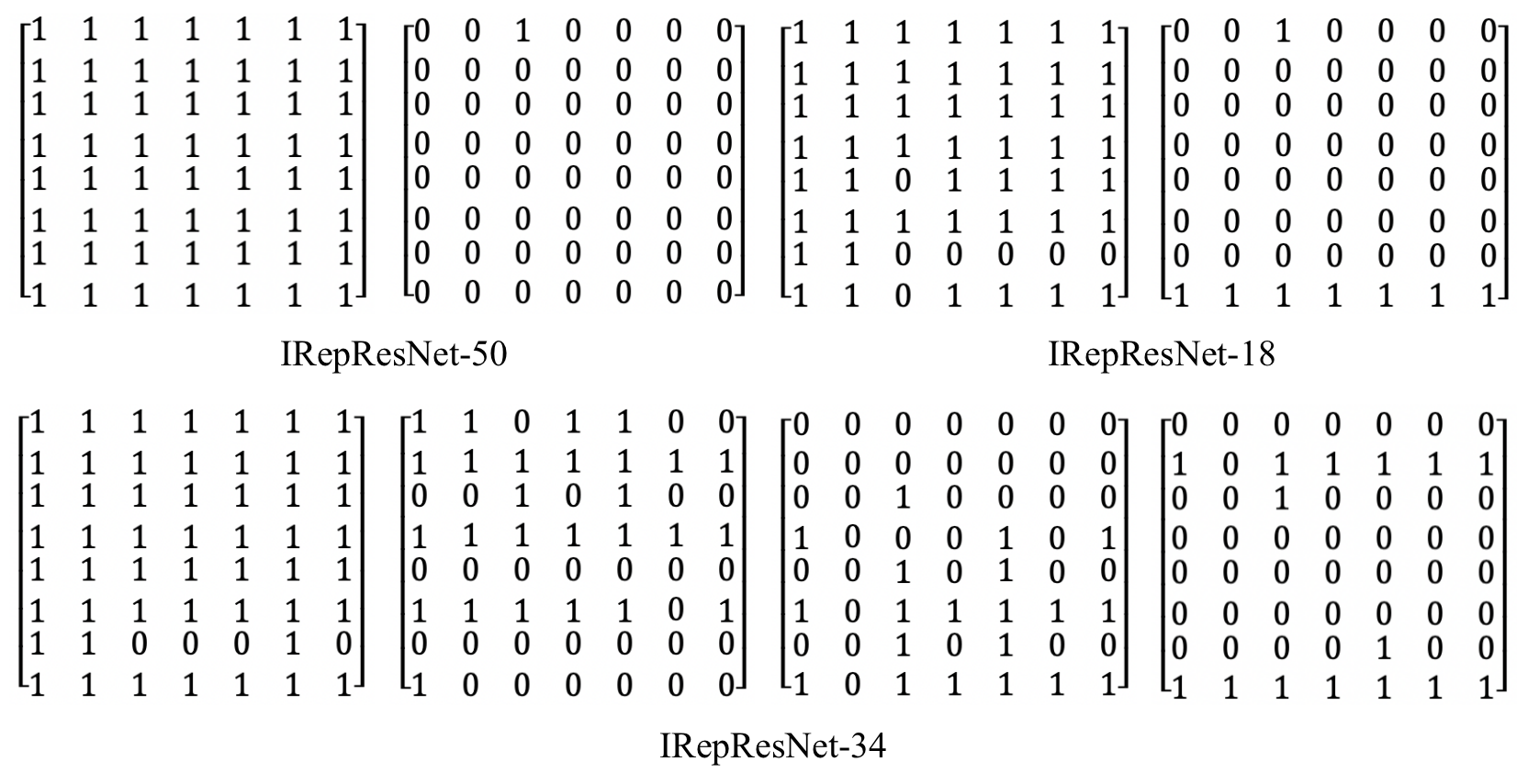}
		\caption{List of architectures of IrepResnet-50,  IrepResnet-18,  IrepResnet-34. Each row represents the enhancement of a 3×3 convolution. From left to right, it represents the re-parameterization structure from the first layer to the last layer.}\label{fig8}
	\end{figure}
	\begin{figure}[H]
		\centering
		\includegraphics[width=0.9\textwidth]{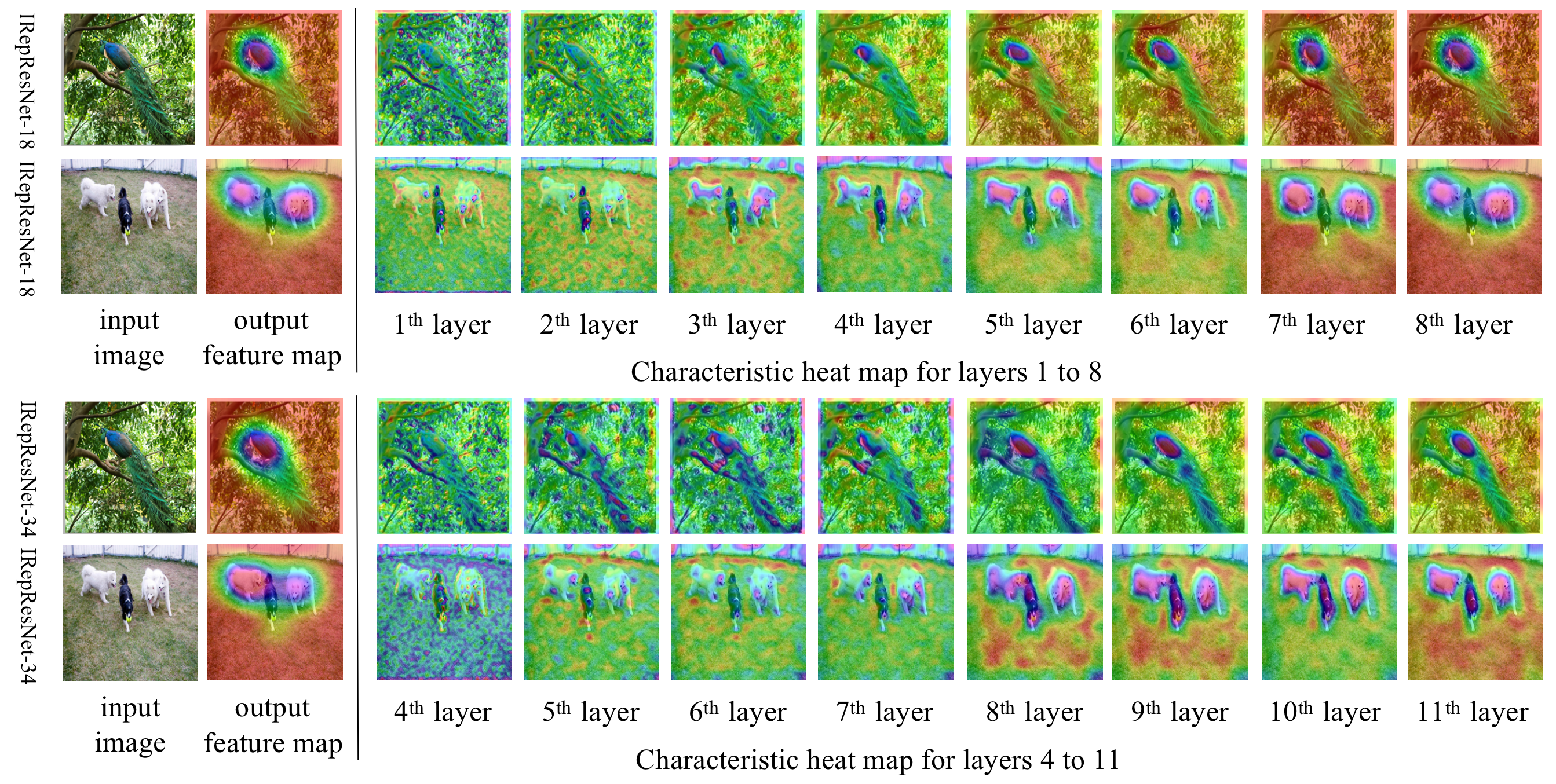}
		\caption{We visualized the output feature values of the convolution in the IrepResNet-18 and  IrepResNet-34 to better interpret the structure of our architecture. We consider the structure of Conv$3\times 3$-Conv$3\times 3$  as one layer.}\label{fig9}
	\end{figure}
\end{appendices}

\bibliography{sn-bibliography}

\end{document}